\documentclass[10pt,twocolumn,letterpaper]{article}

\usepackage{cvpr}
\usepackage{times}
\usepackage{epsfig}
\usepackage{graphicx}
\usepackage{amsmath}
\usepackage{amssymb}

\usepackage{mathtools}
\usepackage{breqn}
\usepackage{verbatim}
\makeatletter
\@namedef{ver@everyshi.sty}{}
\makeatother
\usepackage{array}
\usepackage{subcaption}
\usepackage{enumitem}

\usepackage[pagebackref=true,breaklinks=true,letterpaper=true,colorlinks,bookmarks=false]{hyperref}
\cvprfinalcopy 
\def\cvprPaperID{1127} 

\ifcvprfinal\fi 
\begin{document}

\title{Learning Pyramid-Context Encoder Network for High-Quality Image Inpainting}
\author{Yanhong Zeng$^{1,2}$\thanks{}, Jianlong Fu$^{3}$, Hongyang Chao$^{1,2}$, Baining Guo$^{3}$
	\\ \small $^1$School of Data and Computer Science, Sun Yat-sen University, Guangzhou, P.R. China
	\\ \small $^2$The Key Laboratory of Machine Intelligence and Advanced Computing (Sun Yat-sen University), 
	\\ \small Ministry of Education, Guangzhou, P.R. China
	\\ \small $^3$Microsoft Research, Beijing, P.R. China
	\\ \small zengyh7@mail2.sysu.edu.cn, \{jianf,bainguo\}@microsoft.com, isschhy@mail.sysu.edu.cn}

\twocolumn[{%
	\renewcommand\twocolumn[1][]{#1}%
	\maketitle
	\thispagestyle{empty}
	\begin{center}
		\centering
		\includegraphics[width=\linewidth]{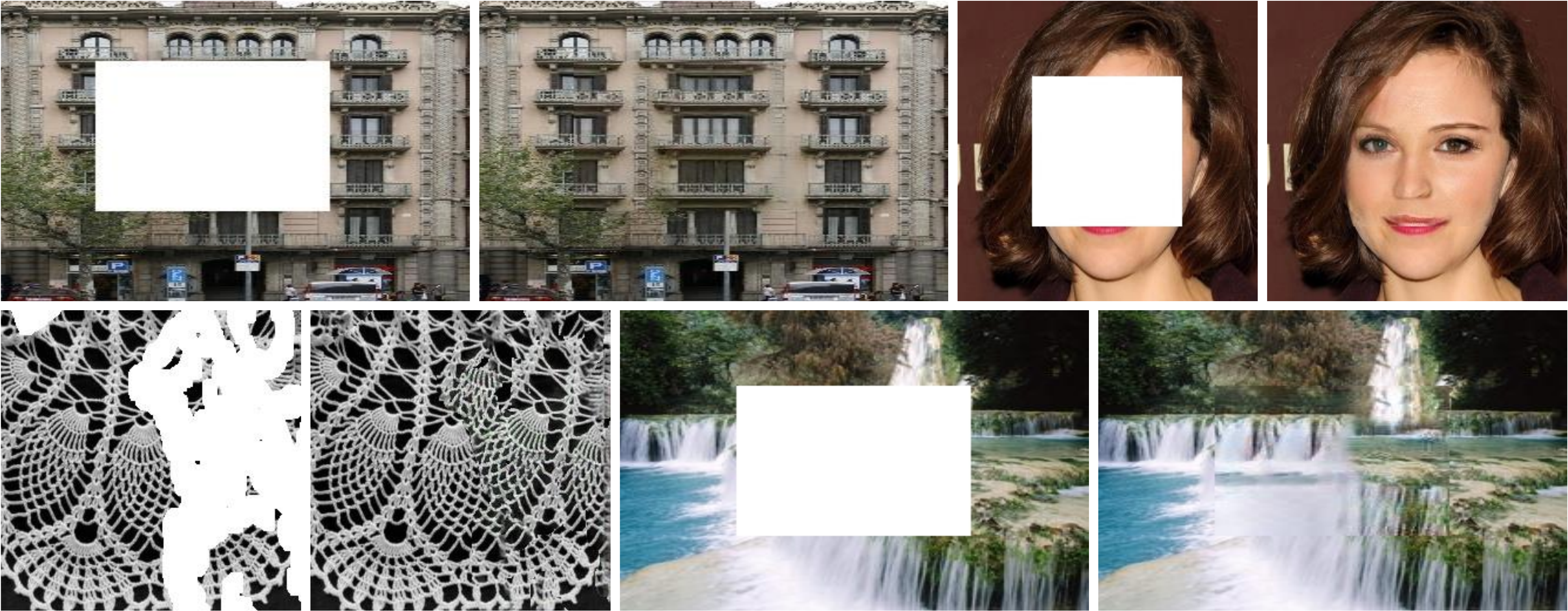}
		\vspace{-5mm}
		\captionof{figure}{High-quality image inpainting results generated by the proposed \textbf{P}yramid-context \textbf{EN}coder \textbf{Net}work (PEN-Net). In each pair, the left is a damaged image masked in white, and the right is the result of image inpainting. PEN-Net shows excellent performance on a variety of images, including facades, natural scene, face and texture. [Best viewed in color]}
		\label{fig:teaser}
	\end{center}%
}]
{
	\renewcommand{\thefootnote}%
	{\fnsymbol{footnote}}
	\footnotetext[1]{This work was performed when the first author was visiting Microsoft Research as a research intern.}
}
\begin{abstract}
	\vspace{-3.5mm}
	High-quality image inpainting requires filling missing regions in a damaged image with plausible content. Existing works either fill the regions by copying image patches or generating semantically-coherent patches from region context, while neglect the fact that both visual and semantic plausibility are highly-demanded. In this paper, we propose a \textbf{P}yramid-context \textbf{EN}coder \textbf{Net}work (PEN-Net) for image inpainting by deep generative models. The PEN-Net is built upon a U-Net structure, which can restore an image by encoding contextual semantics from full resolution input, and decoding the learned semantic features back into images. Specifically, we propose a pyramid-context encoder, which progressively learns region affinity by attention from a high-level semantic feature map and transfers the learned attention to the previous low-level feature map. As the missing content can be filled by attention transfer from deep to shallow in a pyramid fashion, both visual and semantic coherence for image inpainting can be ensured. We further propose a multi-scale decoder with deeply-supervised pyramid losses and an adversarial loss. Such a design not only results in fast convergence in training, but more realistic results in testing. Extensive experiments on various datasets show the superior performance of the proposed network.
\end{abstract}

\vspace{-5mm}
\section{Introduction}
\label{sec:intro}
Image inpainting aims at filling missing pixels in a damaged image given a corresponding mask \cite{Bertalmio2000}. This task has drawn great attention and become a valuable and active research topic for decades \cite{criminisi2004region,levin2004seamless,pathak2016context}, because high-quality image inpainting can benefit a broad range of applications, such as old photo restoration, object removal, and so on.

High-quality image inpainting usually requires synthesizing not only visually-realistic but semantically-reasonable content for missing regions~\cite{bertalmio2003simultaneous, criminisi2004region,yan2018shift,yang2017high, yu2018generative}. Existing approaches can be roughly divided into two groups. As shown in Table~\ref{table:check}, the first group inspired by texture synthesis techniques attempts to fill regions at image-level~\cite{barnes2009patchmatch,criminisi2004region, sun2005image}. Specifically, such approaches usually sample and paste full image resolution patches from source images into missing regions, which allows synthesizing results with details. However, as the lack of high-level understanding of an image, such approaches often fail in generating semantically-reasonable results. To solve this problem, the second group of approaches proposes to encode the semantic context of an image into a latent feature space by deep neural networks and then generate semantic-coherent patches by generative models~\cite{liu2018image, pathak2016context, yu2018generative}. However, it remains challenging to generate visually-realistic results from a compact latent feature, as full image resolution details can be usually smoothed by stacked convolutions and poolings. 

To ensure that both visual and semantic coherence can be satisfied, we propose to fill regions at both image and feature levels. First, we adopt a U-Net~\cite{ronneberger2015u} structure as our backbone, which can encode the context from low-level pixels to high-level semantic features and decode the features back into an image. Specifically, we propose a \textbf{P}yramid-context \textbf{EN}coder \textbf{Net}work (PEN-Net) with three tailored key components, \ie, a pyramid-context encoder, a multi-scale decoder, and an adversarial training loss, to boost the capacity of U-Net in image inpainting. Second, once the compact latent features have been encoded from images, the pyramid-context encoder fills regions from high-level semantic features to low-level features (with richer details) in a pyramid pathway before decoding. To this end, we propose an \textbf{A}ttention \textbf{T}ransfer \textbf{N}etwork (ATN) to learn region affinity between patches inside/outside missing regions in a high-level feature map, and then transfer (\ie, weighted copy by affinity) relevant features from outside into inside regions of previous feature map with higher resolution. Third, the proposed multi-scale decoder takes as input the reconstructed features from ATNs through skip connections and the latent features for final decoding. The PEN-Net is optimized by minimizing deeply-supervised pyramid L1 losses and an adversarial loss. 

To the best of our knowledge, the proposed PEN-Net is the first work that is able to fill missing regions at both image-level and feature-level for image inpainting. we highlight our contributions as follows:
\begin{itemize}[nosep]
	\item \textbf{Cross-layer attention transfer}. We propose a novel network, ATN, to learn region affinity from high-level feature maps (e.g., the compact latent features in the encoder). The resultant affinity map can guide feature transfer in adjacent low-level layers in an encoder.
	\item  \textbf{Pyramid filling}. Our model can fill holes multiple times (depends on the depth of the encoder) by repeating using ATNs from deep to shallow, which can restore an image with more fine-grained details.

\end{itemize}

\begin{table}
	\begin{center}
		\begin{tabular}{m{1.5cm}m{2.4cm}cc}
			\hline
			Category & Method & Semantic  & Details \\
			\hline
			\hline
			image level   &PatchMatch\cite{barnes2009patchmatch}, Region filling\cite{criminisi2004region}         &  &\checkmark \\
			\hline
			feature level &GL\cite{iizuka2017globally}, PConv\cite{liu2018image}, GntIpt\cite{yu2018generative}   &\checkmark  & \\
			\hline
			Ours    & &\checkmark  &\checkmark \\
			\hline
		\end{tabular}
	\end{center}
	\vspace{-5mm}
	\caption{Two groups of typical approaches for image inpainting. PatchMatch \cite{barnes2009patchmatch} and Region filling~\cite{criminisi2004region} ensure that patches with more details can be used for filling, while GL~\cite{iizuka2017globally}, Pconv~\cite{liu2018image} and GntIpt~\cite{yu2018generative} can generate semantic-coherent results. Compared with those methods, our approach can satisfy both semantic and visual requirements.}
	\label{table:check}
\end{table}

\begin{figure*}
	\begin{center}
		\includegraphics[width=\linewidth]{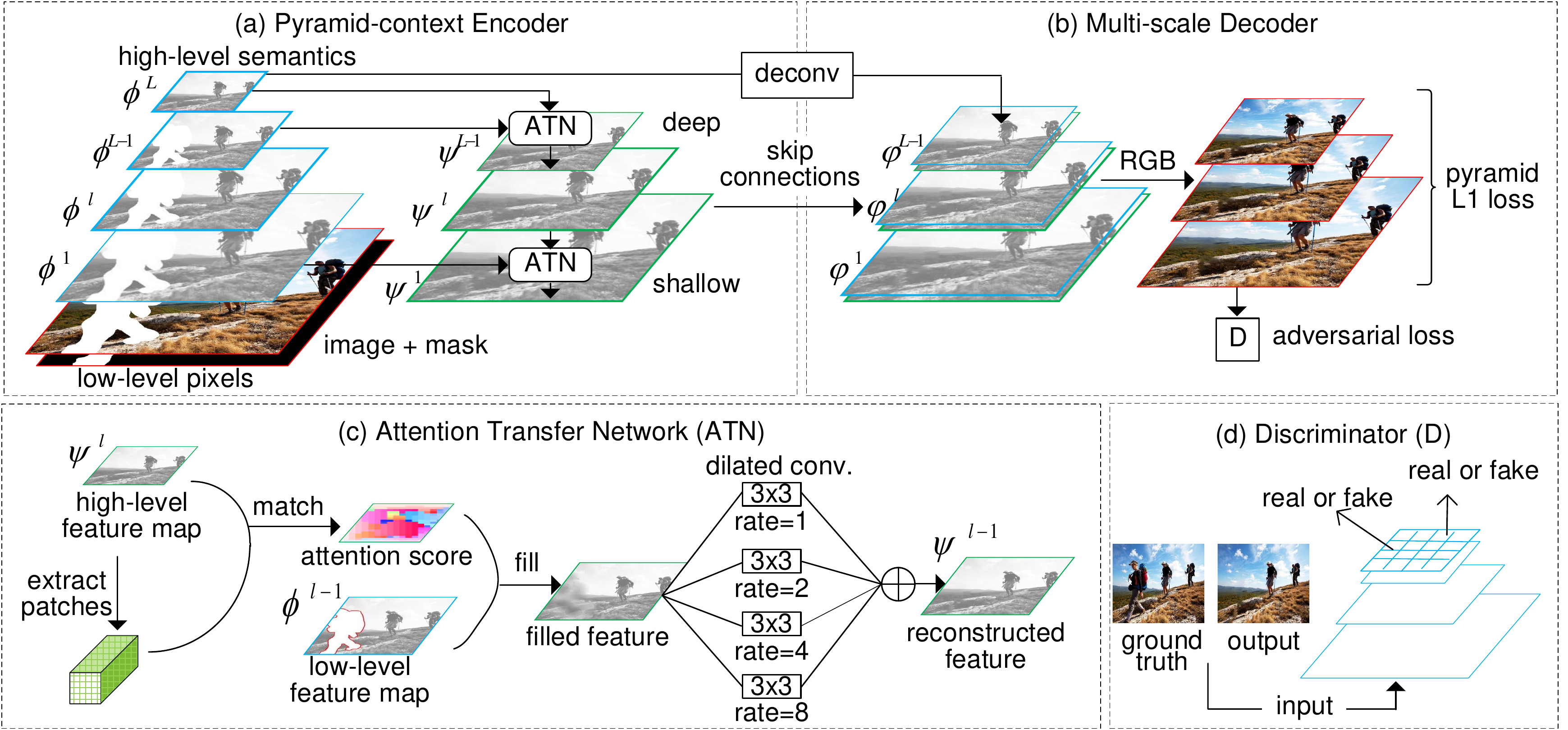} 
	\end{center}
	\vspace{-3mm}
	\caption{The \textbf{P}yramid-context \textbf{En}coder \textbf{Net}work~(PEN-Net) is proposed to boost the capability of U-Net in image inpainting with three tailored components, \ie, a pyramid-context encoder~(a), a multi-scale decoder~(b), and an adversarial training loss~(d). First, once the compact latent feature has been encoded, the pyramid-context encoder further improves the encoding effectiveness by filling regions from high-level feature maps to low-level feature maps (with richer details) through the proposed \textbf{A}ttention \textbf{T}ransfer \textbf{N}etwork~(ATN)~(c). Second, the multi-scale decoder takes as input the reconstructed features from ATNs through skip connections and the latent features for decoding. Finally, the decoder decodes the features back into an image. The whole network is optimized by minimizing pyramid L1 losses and an adversarial loss. [Best viewed in color.]}
	\label{fig:pen}
\end{figure*}

\section{Related Work}
\label{sec:related}
\textbf{Image inpainting by patch-based methods.}~ Patch-based methods were first proposed for texture synthesis~\cite{efros2001image,efros1999texture}. They were then applied in image inpainting to fill missing regions at image level~\cite{telea2004image}. They usually sample and paste similar patches from database or undamaged surroundings into missing regions based on distance metrics between patches~(\eg Euclidean distance, SIFT distance~\cite{lowe1999object}, \etc). Bertalmio \etal proposed to combine patch-based texture synthesis techniques with diffusion-based propagation under image decomposition~\cite{bertalmio2003simultaneous}. A number of approaches try to improve performance by providing better filling order or optimal patches~\cite{criminisi2004region,sun2005image,wexler2007space}. PatchMatch was proposed for quickly finding similar matches between image patches~\cite{barnes2009patchmatch}. Patch-based methods for image inpainting are able to generate sharp results similar with context. However, it's hard to generate semantically-reasonable results by patch-based methods, due to the lack of high level understanding of images.

\textbf{Image inpainting by deep generative models.}~ Deep generative models for image inpainting usually encode an image into a latent feature, fill missing regions at the feature-level, and decode the feature back into an image. Promising results have been achieved by deep generative models recently. Based on deep feature learning and adversarial training, Context Encoder, one of the first deep generative models, is able to give reasonable results for semantic hole-filling~\cite{pathak2016context}. Guidance loss was introduced to make the feature maps generated in decoder as close as possible to the feature maps of ground-truth generated in encoder~\cite{yan2018shift}. Dilated convolutions~\cite{yu2016multi} were introduced to increase receptive field in completion network by Iizuka \etal ~\cite{iizuka2017globally}. Special convolution operations such as PConv~\cite{liu2018image} and ShCNN~\cite{ren2015shepard} were designed for eliminating the effects caused by the placeholder values in masked regions in an image. Contextual attention layer~\cite{yu2018generative} and Patch-swap layer~\cite{song2018contextual} were proposed for filling missing pixels with similar patches from undamaged regions at high-level feature maps. Inspired by image stylization, MNPS proposed to optimize texture details by using a pre-trained classification network during inference~\cite{yang2017high}. Isola \etal try to solve image inpainting by a general image translation framework~\cite{pix2pix2017}. Leveraging high-level semantic feature learning, deep generative models are able to generate semantically-coherent results for the missing regions. However, it remains challenging to generate visually-realistic results from a compact latent feature. 

\section{Pyramid-context Encoder Network}
\label{sec:pen} 
The Pyramid-context Encoder Network~(PEN-Net) consists of three parts (as shown in Figure~\ref{fig:pen}), \ie, a pyramid-context encoder~(a), a multi-scale decoder~(b) and a discriminator~(d). The PEN-Net is built upon a U-Net structure, which can encode a damaged image with mask from full input resolution pixels into a compact latent features and decode the features back into an image. 

As the compact latent features encode the semantics of the context, the pyramid-context encoder can further improves the encoding effectiveness by filling missing regions from the compact latent feature to low-level features (with higher resolution and richer details). It fills holes by repeating using the proposed \textbf{A}ttention \textbf{T}ransfer \textbf{N}etwork~(ATN)~(c) multiple times (according to the depth of the encoder) before decoding. Specifically, an ATN learns region affinity between patches inside/outside missing regions from high-level semantic features, and the learned attention is transferred to fill regions (\ie, weighted copy from the context by affinity) in its previous feature map with higher resolution. Multi-scale information is further aggregated to refine the filled features by four groups of dilated convolutions with different rates in an ATN. Finally, the multi-scale decoder takes as input the reconstructed features from ATNs through skip connections and the latent features for decoding. In addition to an adversarial loss, pyramid L1 losses are used to progressively refine the prediction output by the decoder at all scales. 

We describe details of the pyramid-context encoder and the ATN in Section~\ref{subsec:encoder}. The multi-scale decoder and pyramid L1 losses are introduced in Section~\ref{subsec:decoder} followed by adversarial training loss described in Section~\ref{subsec:dis}. 


\subsection{Pyramid-context encoder}
\label{subsec:encoder}
\textbf{Pyramid-context encoder}~ In order to improve the effectiveness of encoding, the pyramid-context encoder is proposed for filling missing regions before decoding. Once a compact latent feature is learned, the pyramid-context encoder fills regions from high-level semantic features to low-level features (with higher resolution) by repeating using the proposed ATNs in a pyramid fashion. Under the assumption that pixels with similar semantics should have similar details, an ATN is applied at each level to learn region affinity from high-level semantic features, thus the learned region affinity can further guide feature transfer inside/outside missing regions in an adjacent layer with higher resolution. 

Given a pyramid-context encoder of $L$ layers, we denote the feature maps from deep to shallow as $\phi^L, \phi^{L-1}, ..., \phi^1$ as shown in (a) of Figure~\ref{fig:pen}. 
The features constructed by ATNs in each layer from deep to shallow are denoted as:
\begin{gather}
\psi^{L-1} = f(\phi^{L-1},  \phi^{L}), \nonumber \\
\psi^{L-2} = f(\phi^{L-2},  \psi^{L-1}),  \\
\cdots, \nonumber \\ 
\psi^{1} = f(\phi^1, \psi^2) = f(\phi^1, f(\phi^2, ...f(\phi^{L-1},  \phi^{L}))),\nonumber
\end{gather}
\label{eq:H}where we denote the operation of the ATN as $f$. By such a cross-layer attention transfer and pyramid filling mechanism, both visual and semantic coherence for the missing regions can be ensured. The details of $f$ (\ie, ATN) are introduced as below. 

\textbf{Attention Transfer Network}~ 
We follow state-of-the-art approaches to fill missing regions by using attention \cite{song2018contextual, yan2018shift, yu2018generative}. The attention is usually obtained by region affinity between patches (usually $3\times3$) inside/outside missing regions, thus relevant features outside can be transferred (\ie, weighted copy from the context by affinity) into inside regions. As shown in (c) of Figure~\ref{fig:pen}, the ATN first learns region affinity from a high-level feature map, $\psi^l$. It extracts patches from $\psi^l$ and calculate the cosine similarity between patches inside and outside missing regions:
\begin{equation}
s^l_{i,j}=\langle \frac{p^l_i}{\left \|p^l_i\right\|_2}, \frac{p^l_j}{\left\|p^l_j\right\|_2} \rangle
\label{eq: similarity},
\end{equation}
where $p_i^l$ is the $i$-th patch extracted from $\psi^l$ outside mask, $p_j^l$ is the $j$-th patch extracted from $\psi^l$ inside the mask. Then softmax is applied on the similarities to obtain the attention score for each patch:
\begin{equation}
\alpha_{j,i}^l=\frac{\exp (s^l_{i,j})}{\sum_{i=1}^{N}\exp(s^l_{i,j})}
\label{eq:softmax}.
\end{equation}
After obtaining the attention score from a high-level feature map, the holes in its adjacent low-level feature map can be filled with context weighted by the attention score:
\begin{equation}
p_j^{l-1} = \sum_{i=1}^{N}\alpha_{j,i}^l\ \  p^{l-1}_{i}
\label{eq:out},
\end{equation}
where $p_i^{l-1}$ is the $i$-th patch extracted from $\phi^{l-1}$ outside masked regions, and $p_j^{l-1}$ is the $j$-th patch to be filled in missing regions. After calculating all patches, we can finally obtain a filled feature $\psi^{l-1}$ by attention transfer from $\psi^l$. In particular, all these operations can be formulated into convolution operations for end-to-end training \cite{yu2018generative}.

We propose to further refine the filled features in an ATN as shown in (c) of Figure~\ref{fig:pen}. Specifically, multi-scale contextual information can be aggregated by four groups of dilated convolutions with different rates. Such a design ensures structure coherence with context in the final reconstructed features, which improves the inpainting results in testing. 

\begin{figure*}
	\begin{center}
		\includegraphics[width=\linewidth]{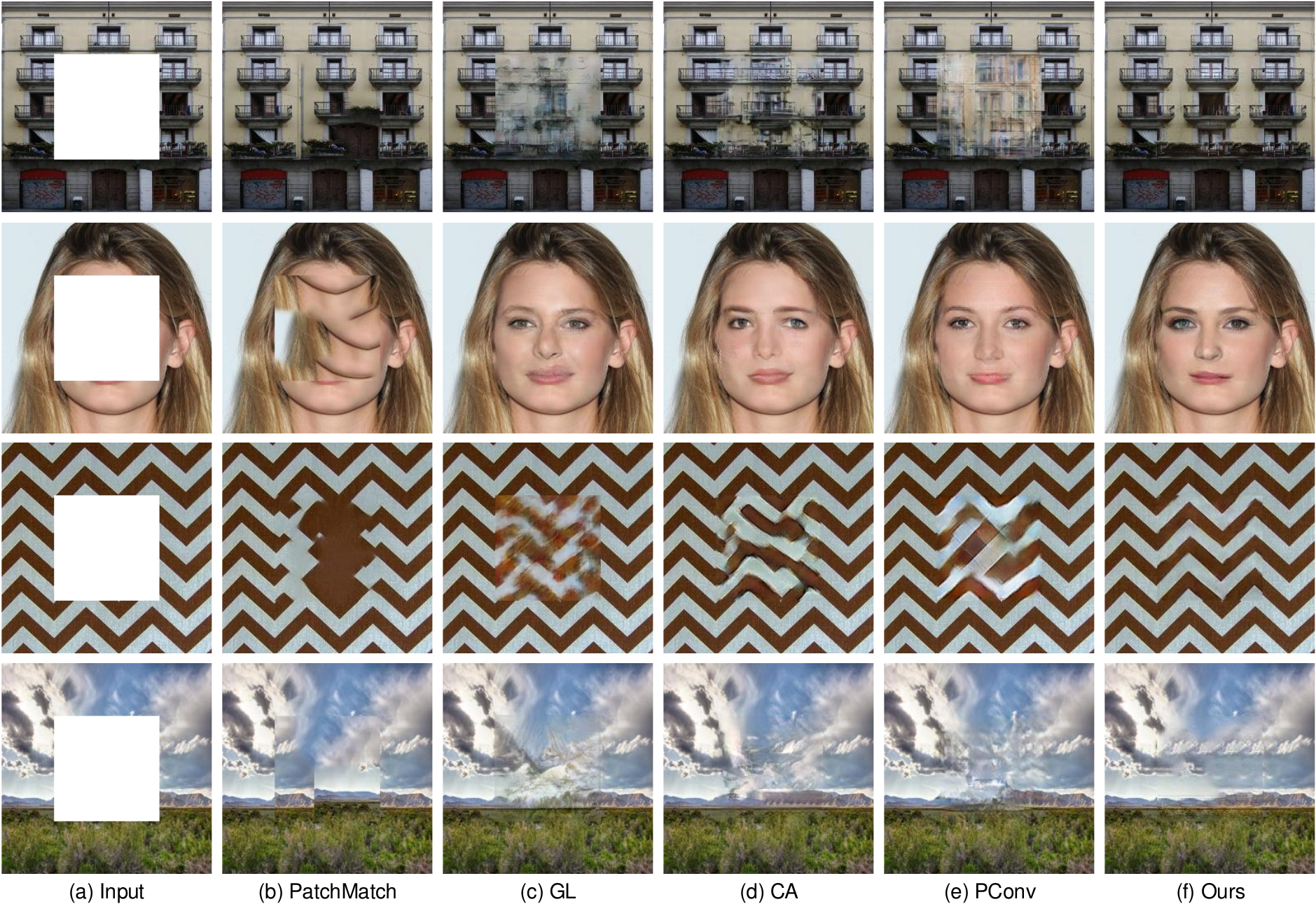}
	\end{center}
    \vspace{-5mm}
	\caption{Qualitative comparisons with baselines on four datasets with different characteristics. In each row, the first image is the input with a large mask in the center~(\ie, $128 \times 128$), and the left images from left to right are the results generated by PatchMatch \cite{barnes2009patchmatch}, GL~\cite{iizuka2017globally}, CA~\cite{yu2018generative}, PConv~\cite{liu2018image} and our model respectively.~[Best viewed with zoom-in.] }
	\label{fig:exp-toy}
\end{figure*}

\subsection{Multi-scale decoder}
\label{subsec:decoder}

\textbf{Multi-scale decoder}~ The proposed multi-scale decoder takes as input the reconstructed features from ATNs through skip connections and the latent features from the encoder. We denote the feature maps generated by the multi-scale decoder as $\varphi^{L-1}, \varphi^{L-2}, ..., \varphi^{1}$ from deep to shallow, which are obtained as follows:
\begin{gather}
\varphi^{L-1} = \psi^{L-1} \oplus g(\phi^L), \nonumber \\
\varphi^{L-2} = \psi^{L-2} \oplus g(\varphi^{L-1}),  \\
\cdots, \nonumber \\
\varphi^{1} = \psi^{1} \oplus g(\varphi^{2}), \nonumber
\end{gather}
\label{eq:D}where $g$ denotes transposed convolution operation, $\oplus$ denotes feature concatenation, and $\psi^l$ is the reconstructed feature from an ATN in the $l$-th layer of the encoder. 

On one hand, the reconstructed features generated by ATNs encode more low-level information for missing regions. Such a design enables the decoder to generate visually realistic results with fine-grained details. On the other hand, the features obtained from the compact latent features by convolutions are able to synthesize novel objects in missing regions, even when the objects cannot be found outside missing regions. Combining those two kinds of features, the decoder is able to synthesize novel objects with high coherence in semantics and textures with the context of the image. For example, the proposed decoder is able to synthesize eyes in human face images with both eyes masked. 

\textbf{Pyramid L1 losses}~ We also propose deeply-supervised pyramid L1 losses to progressively refine the predictions for missing regions at each scale. Specifically, each pyramid loss is a normalized L1 distance between a prediction of specific scale and the corresponding ground truth:
\begin{equation}
L_{pd} = \sum_{l=1}^{L-1} \left \|x^l - h(\varphi^{l}) \right\|_1
\label{eq:pd_loss},
\end{equation}
where $h$ denotes a $1\times 1$ convolution which decodes $\varphi^l$ into an RGB image with the same size, and $x^l$ is the ground truth scaled to the same size as $\varphi^l$. The overall objective function incorporating pyramid L1 losses and an adversarial loss is described in the next section. 

\subsection{Adversarial training loss}
\label{subsec:dis}
As image inpainting is an ill-posed problem that there are many possible results for the missing regions, we use adversarial training to select the most realistic one. Adversarial training usually involves a generator (G) and a discriminator (D), which aims at achieving a Nash equilibrium, so that fake data generated by the generator cannot be distinguished from real data by the discriminator. As shown in (d) of Figure~\ref{fig:pen}, the pyramid-context encoder and the multi-scale decoder form a generator, and we adopt PatchGAN~\cite{pix2pix2017} as our discriminator. Spectral normalization is used in the discriminator to stabilize the training \cite{miyato2018spectral}.

We first define the final prediction from the generator as:
\begin{equation}
z = G(x \odot(1-M), M)\odot M + x\odot(1-M)
\label{eq:z},
\end{equation}
where $x$ is the ground truth, $\odot$ is an element-wise multiplication, $M$ is the mask where 1 labels missing regions and 0 labels context. The hinge version of the adversarial loss for the discriminator can be denoted as:
\begin{dmath}
	L_{D}= \mathbb{E}_{x \sim p_{data}} \left[max(0, 1-D(x))\right] + \mathbb{E}_{z \sim p_z}\left[max(0,1+D(z))\right]
	\label{eq:d_loss},
\end{dmath}
where $D(x)$ and $D(z)$ are the logits output from $D$. The adversarial loss for the generator can be denoted as: 
\begin{equation}
L_{G} = - \mathbb{E}_{z \sim p_z}\left[D(z)\right]
\label{eq:g_loss}.
\end{equation}
The whole PEN-Net is optimized by minimizing an adversarial loss and pyramid L1 losses defined in Section~\ref{subsec:decoder}. We define the overall objective function as:
\begin{equation}
L = \lambda_{G}L_G + \lambda_{pd}L_{pd}
\label{eq:joint_loss}.
\end{equation}

\section{Experiments}
\label{sec:exp}
We evaluate the proposed network with baselines from both quantitative and qualitative aspects. Details of experimental settings are introduced in Section~\ref{subsec:base}, and the experiments results are described in Section~\ref{subsec:res}, followed by the analysis of the effectiveness of our model in Section~\ref{subsec:analysis}.

\subsection{Experimental settings}
\label{subsec:base}
\textbf{Datasets}~ We conduct experiments on four datasets with different characteristics as below (details in Table~\ref{table:dataset}):
\begin{itemize}[nosep]
	\item[--] Facade~\cite{Tylecek13}: a collection of highly-structured facades from different cities around the world.
	\item[--] DTD~\cite{cimpoi14describing}, an evolving dataset of 47 kinds of describable textures collected in the wild. 
	\item[--] CELEBA-HQ~\cite{karras2017progressive}, a high-quality version of the human face dataset from CELEBA~\cite{liu2015faceattributes}. 
	\item[--] Places2~\cite{zhou2018places}, a dataset that contains images of 365 scenes collected from the natural world.  
\end{itemize}

\textbf{Baselines}~ We compare with the following baselines for their state-of-the-art performance: 
\begin{itemize}[nosep]
	\item[--] PatchMatch (PM) \cite{barnes2009patchmatch}: a typical patch-based approach, which copies similar patches from the surroundings.
	\item[--] GL~\cite{iizuka2017globally}: a generative model, which leverages both global and local discriminators for image completion. 
	\item[--] CA~\cite{yu2018generative}: a two-stage inpainting model, which leverages contextual attention at high-level features. 
	\item[--] PConv~\cite{liu2018image}: a generative model, which proposes a special convolution layer for filling irregular holes.
\end{itemize}

\textbf{Implementation details} 
We use random blocks for training, following the experimental settings used by baselines \cite{iizuka2017globally,yu2018generative} for fair comparisons. All images are resized to $256\times256$ for training and testing. When extracting hole and non-hole patches in each level, we use nearest neighbor down-sampling to evolve the holes. Our full model runs at 0.19 seconds per frame on a GPU TITAN V for images of size $256\times256$. All the results reported are output directly from the trained models without using any post-processing. The code will be made publicly available. \footnote{https://github.com/researchmm/PEN-Net-for-Inpainting}

\subsection{Results}
\label{subsec:res}

\textbf{Quantitative comparisons}~ As Places2 contains natural-world images, which is considered as the most challenging dataset \cite{iizuka2017globally,yu2018generative} (compared with Facade/DTD/CELEBA-HQ), we conduct quantitative comparisons on Places2. All images are randomly masked with $128\times128$ squares for testing. We use L1 loss, multi-scale structural similarity (MS-SSIM) \cite{wang2003multiscale}, Inception Score (IS) \cite{salimans2016improved} and Fr\'{e}chet Inception Distance (FID) \cite{heusel2017gans} as evaluation metrics. The results listed in Table~\ref{table:exp_num} show the comparable performance of the proposed approach against baselines. 

L1 loss can roughly reflect the ability of models to reconstruct the original image content. MS-SSIM extracts and evaluates the similarity of structural information from paired images in multi-scale to provide a good approximation to human visual perception. However, there are a great deal of solutions different from original content for the missing regions, while L1 loss and MS-SSIM are limited to comparing with the original image content. Under the assumption that a damaged scene image should maintain the same attributes after image inpainting, an inpainting result should be confidently identified as a specific category by the pre-trained classification network. To this end, we also use the inception score as one of the evaluation metrics:
\begin{equation}
I = \exp (  \mathop{\mathbb{E}}_{z \sim p_z} [(D_{KL}(p(y|z))\| p(y))] ),
\end{equation}
where $z$ is inpainting results defined in Section~\ref{subsec:decoder}, and $y$ is the label predicted by pre-trained classification models. We use the pre-trained classification network released by Zhou \etal~\cite{zhou2018places}. Besides, FID has driven an increasing attention and becomes a commonly-used numeric metric in the field of image generation. We also include FID to measure the Wasserstein-2 distance between real and fake images using a pre-trained Inception-V3 model \cite{szegedy2016rethinking}. 

\begin{table}
	\begin{center}
		\begin{tabular}{cccc}
			\hline
			Dataset &\#Train &\#Test\ &\#Total \\
			\hline
			\hline
			Facade~\cite{Tylecek13} & 506 & 100  &606 \\
			\hline
			DTD~\cite{cimpoi14describing}  & 4,512 & 1,128 &5,560 \\
			\hline
			CELEBA-HQ~\cite{karras2017progressive} &28,000 &2,000 & 30,000\\
			\hline 
			Places2~\cite{zhou2018places}  &1,803,460   &36,500 & 1,839,960 \\
			\hline
		\end{tabular}
	\end{center}
		\vspace{-6mm}
	\caption{Training and test splits of four datasets.}
	\label{table:dataset}
\end{table}

\begin{table}
	\small
	\begin{center}
	\begin{tabular}{lcccc}
		\hline
		Method        & L1 Loss$\dag$   & MS-SSIM$\P$ & IS$\P$       & FID$\dag$\\ 
		\hline
		\hline
		PatchMatch \cite{barnes2009patchmatch}   &12.90    &60.00\%      &43.03     &20.36 \\
		GL~\cite{iizuka2017globally}                 &9.27	 &73.40\%	        &42.05     &19.18 \\
		PConv~\cite{liu2018image}    	    &\textbf{8.92}	  &74.67\%	        &47.00     &18.39\\
		\hline
		CA~\cite{yu2018generative}    &9.91	    &73.02\%	     &44.81      &18.34 \\
		PEN-Net (ours)  &9.94 &\textbf{78.09\%} &\textbf{50.51}  &\textbf{15.19}\\
		\hline
	\end{tabular}
\end{center}
\vspace{-6mm}
	\caption{Quantitative comparisons on Places2 with L1 Loss, MS-SSIM, IS and FID. $\dag$ Lower is better. $\P$ Higher is better.}
	\label{table:exp_num}
\end{table}

\begin{figure*}
	\begin{center}
		\includegraphics[width=\linewidth]{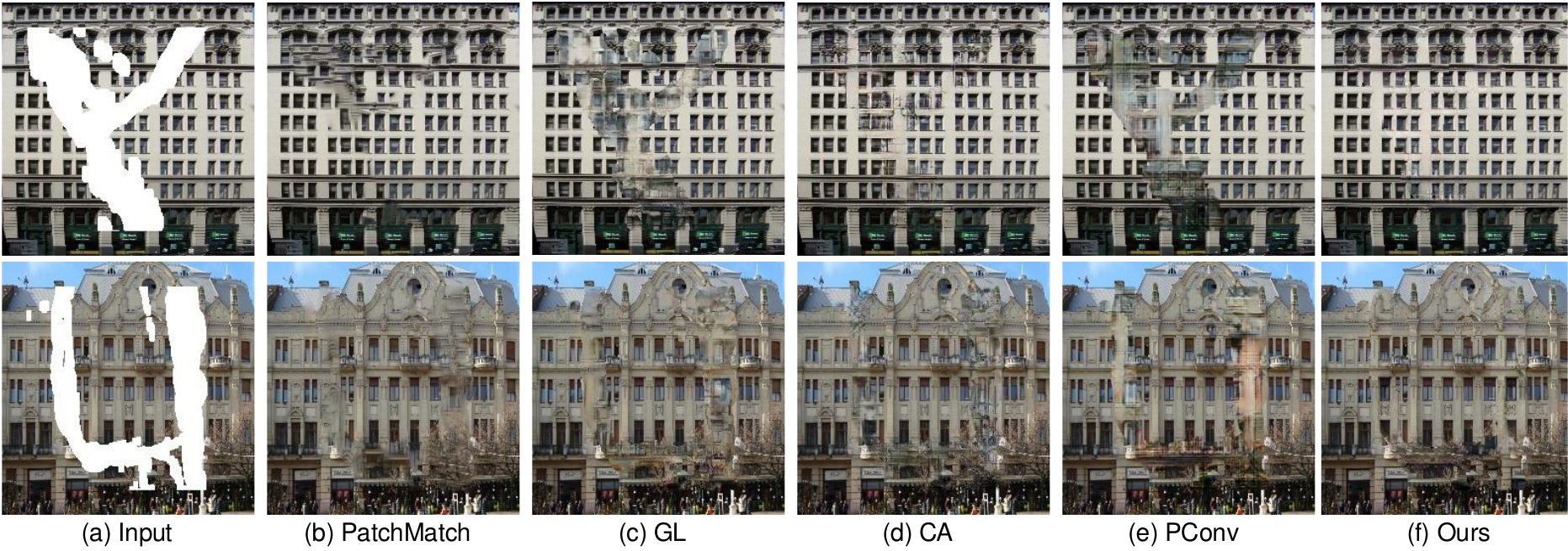}
	\end{center}
	\vspace{-5mm}
	\caption{Qualitative comparisons for image inpainting with irregular masks on Facade.~[Best viewed with zoom-in.]}
	\label{fig:exp-ir}
\end{figure*}

\begin{figure}
	\begin{center}
		\includegraphics[width=\linewidth]{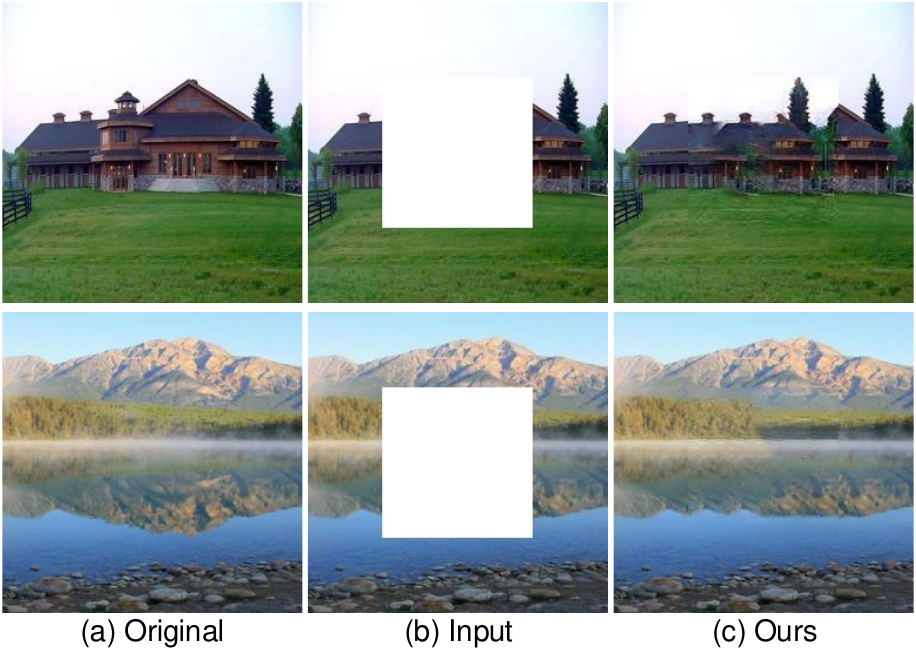}
	\end{center}
			\vspace{-5mm}
	\caption{Example results generated by the proposed network on Places2.~[Best viewed with zoom-in.]}
	\label{fig:exp-others}
\end{figure}

\begin{figure}
	\begin{center}
		\includegraphics[width=\linewidth]{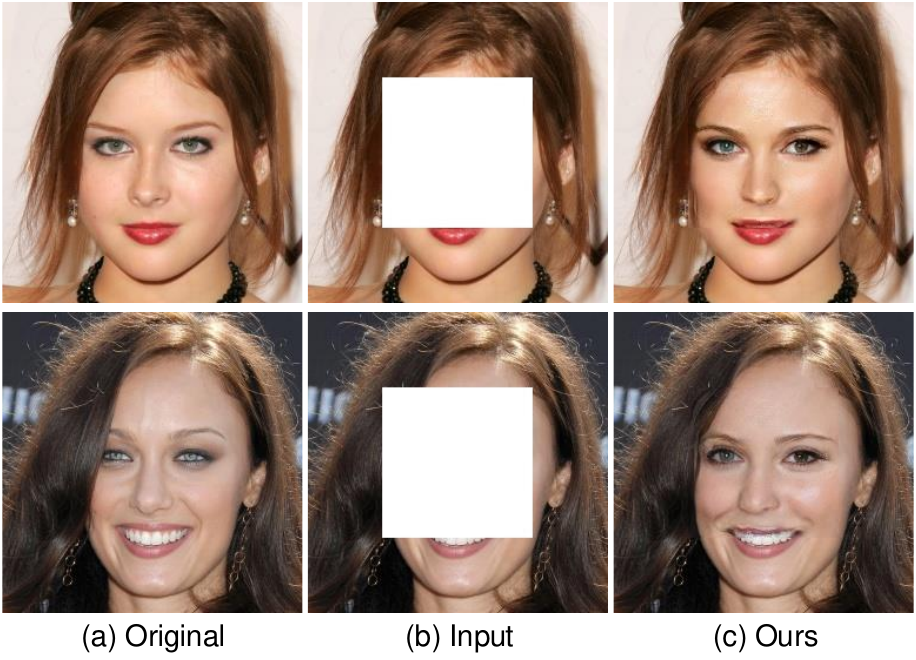}
	\end{center}
				\vspace{-5mm}
	\caption{Example results generated by the proposed network on CELEBA-HQ.~[Best viewed with zoom-in.]}
	\label{fig:exp-face}
\end{figure}

\textbf{Qualitative comparisons}~ In order to take both visual and semantic coherence into account, we conduct qualitative comparisons on the test set of four datasets with different characteristics, which are highly-structured with fine-grained textures. We masked the test images with center $128\times128$ squares, and our model shows superior performance against the state-of-the-art. As shown in Figure~\ref{fig:exp-toy}, the typical patch-based method, PatchMatch, is able to generate clear textures but with distorted structures inconsistent with surrounding areas, while deep generative models including GL, CA and PConv tend to generate blurry textures in the final results. With the help of cross-layer attention transfer and pyramid filling mechanisms, our model is able to generate semantically-reasonable and visually-realistic results with clear textures and consistent structures with context. We also verify the ability of the proposed network to fill missing regions given irregular masks. Specifically, we use the images of Facade and masks released by Liu \etal~\cite{liu2018image} for testing. As shown in Figure~\ref{fig:exp-ir}, the baselines tend to create color discrepancies and distorted structures, while our model outperforms the-state-of-the-art with consistent colors and structures. More example results generated by our model on images of natural scene and human face can be found in Figure~\ref{fig:exp-others} and Figure~\ref{fig:exp-face}.

\textbf{User study}~ In addition to quantitative and qualitative comparisons, we also perform two settings of user study, \ie, paired images and single image user study. The volunteers are all image experts with image processing background. They are not informed of mask information.

In the first setting, over 20 volunteers are invited to evaluate the performance of the models on Facade. Each time, a pair of images generated from different models are shown to the volunteers in an anonymous way. The volunteers are asked to choose the more natural one from those two images. We collected 613 valid votes in total, and the statistics of the results are shown in Table~\ref{table:exp_u1}. The statistics show that our model is ranked better in most of time~(82.10\%) over other models. We also found that people prefer clear results generated by PatchMatch (PM), CA and ours. 

In the second setting, we randomly distribute the validation set of DTD into four groups. Images in three groups are masked with $32\times 32$, $64\times 64$ or $128\times 128$ squares, and the last group is unmasked. Over 25 volunteers are invited to evaluate the naturalness of inpainting results generated by our model with different mask size. Each time, an image sampled from real data or our inpainting results is shown to the volunteers to guess whether the image is a real image from the dataset. We collected 1,425 valid votes in total, and the statistics are shown in Table~\ref{table:exp_u2}. We found that, the inpainting results from the group with $32\times32$ masks can be considered as real in 82.23\% of the time. Even in the challenging $128\times128$ case, we received $32.70\%$ votes.  

\begin{table}
	\small
	\begin{center}
		\begin{tabular}{l|ccccc}
			\hline
			Method & PM &GL &CA &PConv &Ours\\ 
			\hline
			Percentage & 40.15\%  &34.25\% &70.30\% &23.70\% &82.10\%\\
			\hline
		\end{tabular}
	\end{center}
	\vspace{-3mm}
	\caption{Statistics of paired images user study. The value indicates the percentage of being ranked as better.}
	\label{table:exp_u1}
\end{table}

\begin{table}
	\begin{center}
		\begin{tabular}{l|cccc}
			\hline
			Mask size &0~(real) &32 &64 &128\\ 
			\hline
			Percentage & 92.66\%  &82.23\%  &52.63\%  &32.70\%\\
			\hline
		\end{tabular}
	\end{center}
	\vspace{-3mm}
	\caption{Statistics of single image user study. The value indicates the percentage of being considered as real. }
	\label{table:exp_u2} 
\end{table}

\begin{table}
	\begin{center}
		\begin{tabular}{lcccc}
			\hline
			Method & L1 loss$\dag$ & ms-ssim$\P$ &IS$\P$  &FID$\dag$\\
			\hline
			\hline
			patch-swap \cite{song2018contextual}     &12.13   &64.00\%   &29.26    &36.85 \\
			single ATN~(ours)             &\textbf{9.85}    &71.61\%  &37.02    &26.38\\
			PEN-Net~(ours)         &9.94	   &\textbf{78.09\%} &\textbf{50.51}  &\textbf{15.19}\\
			\hline
		\end{tabular}
	\end{center}
	\vspace{-3mm}
	\caption{Ablation comparison of cross-layer attention transfer network (ATN) and pyramid filling mechanism over Places2. $\dag$ Lower is better. $\P$ Higher is better.}
	\label{t2}
	\vspace{-3mm}
\end{table}
	
\subsection{Analysis}
\label{subsec:analysis}
We analyze the effectiveness of different components of the proposed network by visualizing the learned feature maps or ablation study as follows. 

\begin{figure}
	\begin{center}
		\includegraphics[width=\linewidth]{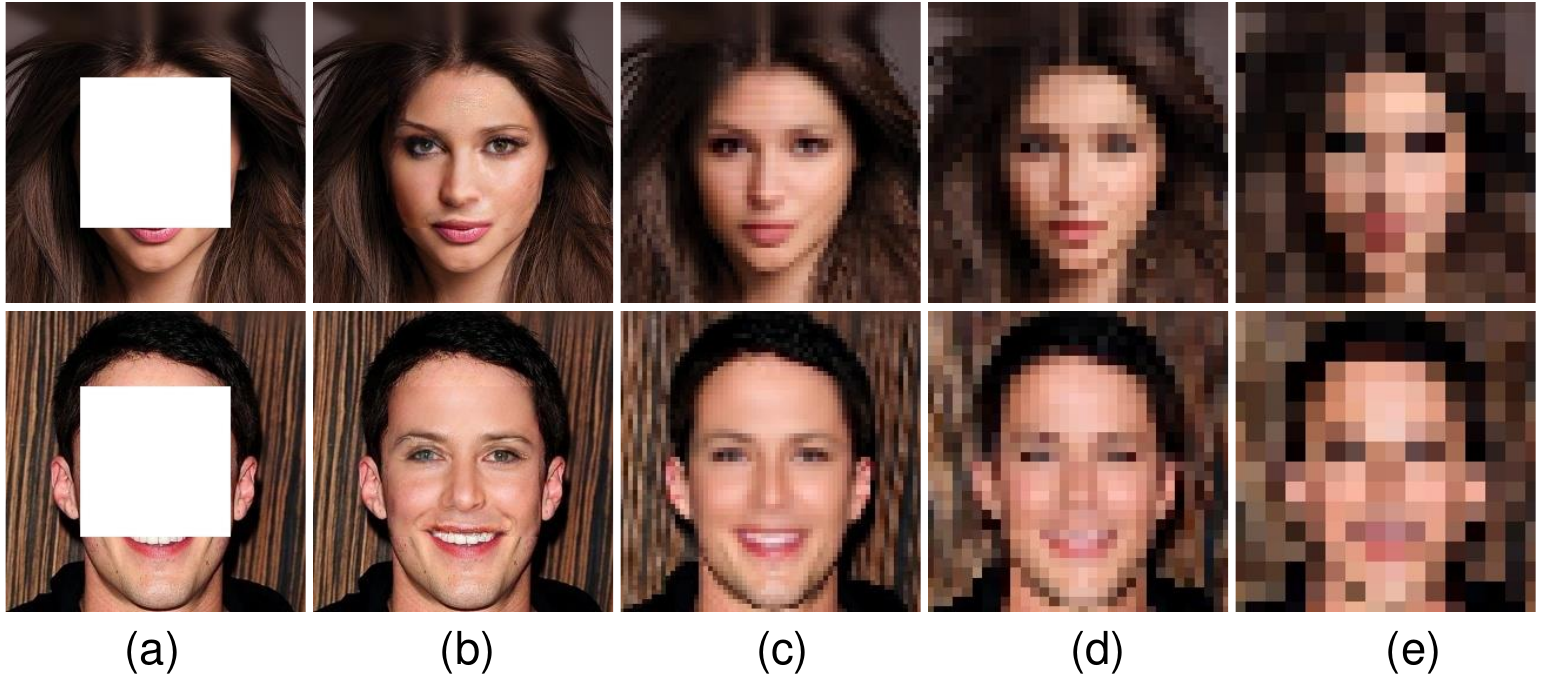}
	\end{center}
	\vspace{-5mm}
	\caption{Images generated by the decoder at each scale. (a) is the input. (b) is the final prediction generated by our model. (c), (d) and (e) are prediction output by the decoder at multiple scales~(all resized to $256\times 256$ for visualization).~[Best viewed with zoom-in.]}
	\label{fig:ab-pdloss}
\end{figure}

\textbf{Effectiveness of the pyramid L1 loss}~ Pyramid L1 losses is proposed to progressively refine the predictions at each scale. We conduct experiments on images of human faces and visualize the images decoded at each scale. As shown in Figure~\ref{fig:ab-pdloss}, the pyramid loss is helpful to decode the compact latent feature into an image layer by layer. 

\begin{figure}
	\begin{center}
		\includegraphics[width=\linewidth]{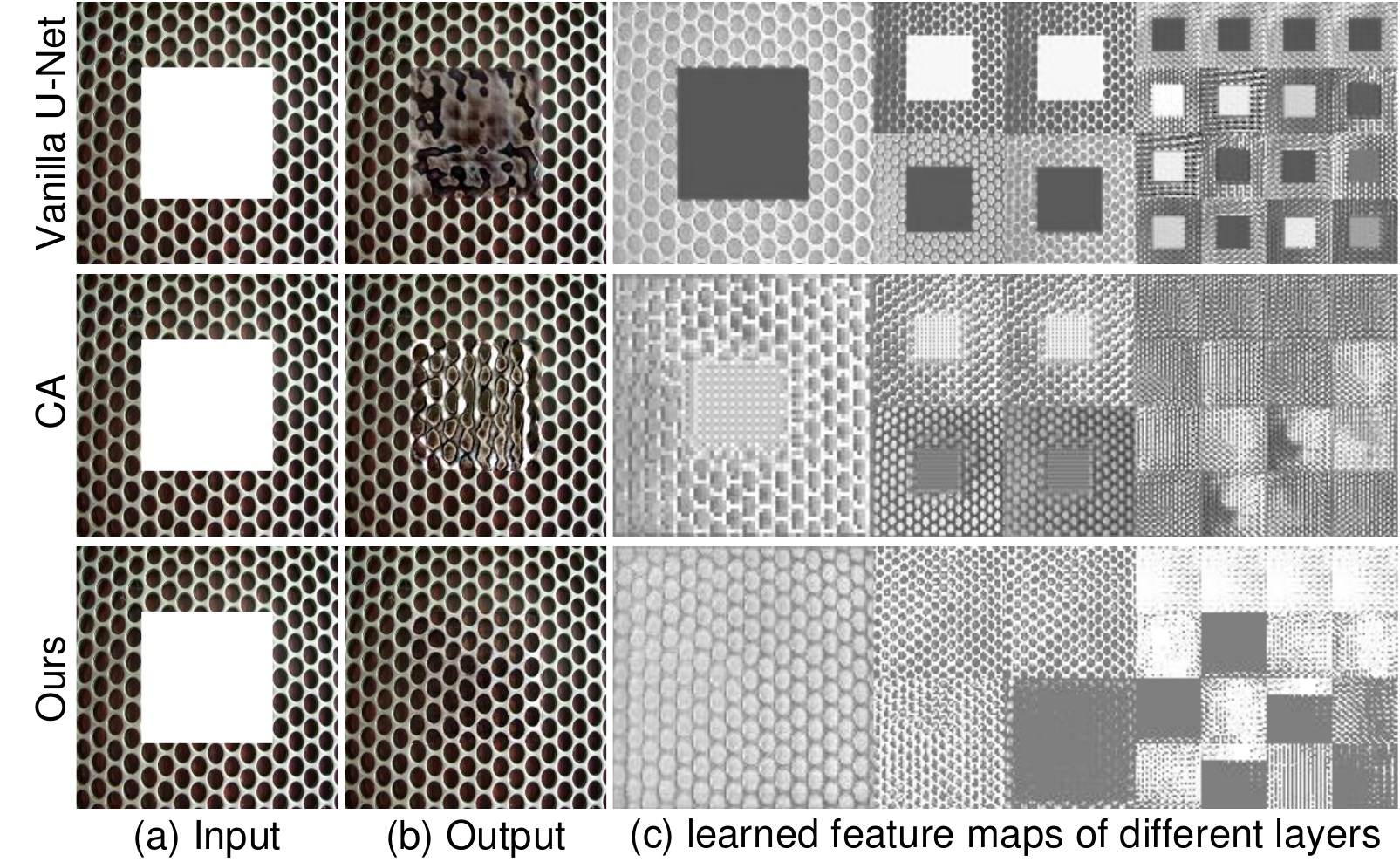}
	\end{center}
	\vspace{-5mm}
	\caption{Visualization of the feature maps learned by the encoder. (a) is the input. (b) is the final prediction generated by the models. (c) are visualized feature maps from different layers.~[Best viewed with zoom-in.]}
	\label{fig:ab-atn}
\end{figure}

\textbf{Effectiveness of the ATN}~
In order to verify the effectiveness of the attention transfer network~(ATN), we visualize the learned feature maps on a same U-Net backbone with different attention mechanisms. As shown in Figure~\ref{fig:ab-atn}, the vanilla U-Net encoder (without using attention) encodes little information inside missing regions, and it fails in generating plausible results. Without the guidance (\ie, attention map) from deeper layers, CA \cite{yu2018generative} (the commonly-used attention method) failed in filling coherent patches inside the missing regions in shallow layers. With the proposed cross-layer ATN, our model is able to fill regions with coherent patches. In addition to comparing with CA in Figure \ref{fig:ab-atn}, we further compare with patch-swap layer \cite{song2018contextual} (the latest attention method) on a same U-Net backbone in Table \ref{t2}. We can observe that, both cross-layer attention transfer network and pyramid filling mechanism bring improvements of performance on a U-Net backbone. 


\section{Conclusion}
\label{sec:conclution}
In this paper, we propose a Pyramid-context Encoder Network~(PEN-Net) to generate semantically-reasonable and visually-realistic results for image inpainting. Specifically, the proposed network boosts both the encoding and decoding effectiveness of a vanilla U-Net by using a pyramid-context encoder and a multi-scale decoder. We highlight two key differences of the attention transfer network used in the encoder, cross-layer attention transfer and pyramid filling from high-level semantic features to low-level features with more details. As a future work, we plan to refine the proposed network for higher resolution images.

\textbf{Acknowledgments} This  work  is  partially  supported by NSF of China under Grant 61672548, U1611461, 61173081, and the Guangzhou Science and Technology Program, China, under Grant 201510010165.

{\small
	\bibliographystyle{ieee}
	\bibliography{egbib}
}

\newpage
\section*{Supplementary Material}
In this section, we present more details of the network architectures and training, additional qualitative comparisons and visual results. We will release the code in the future. 

\addcontentsline{toc}{section}{Appendices}
\renewcommand{\thesubsection}{\Alph{subsection}}

\subsection{Network architectures and training details}
Details of the PEN-Net are listed in Table~\ref{table:encoder}~(SUPP.), Table~\ref{table:decoder}~(SUPP.) and Table~\ref{table:D}~(SUPP.) respectively. We follow the setting of GntIpt, all images are resized to $256\times 256$ with $128\times 128$ square masks in training. We implement GL~\cite{iizuka2017globally} and PConv~\cite{liu2018image} for comparisons, and the performance have achieved the same as reported.

\subsection{More qualitative comparisons and visual results}
\begin{figure*}
	\begin{center}
		\includegraphics[width=\linewidth]{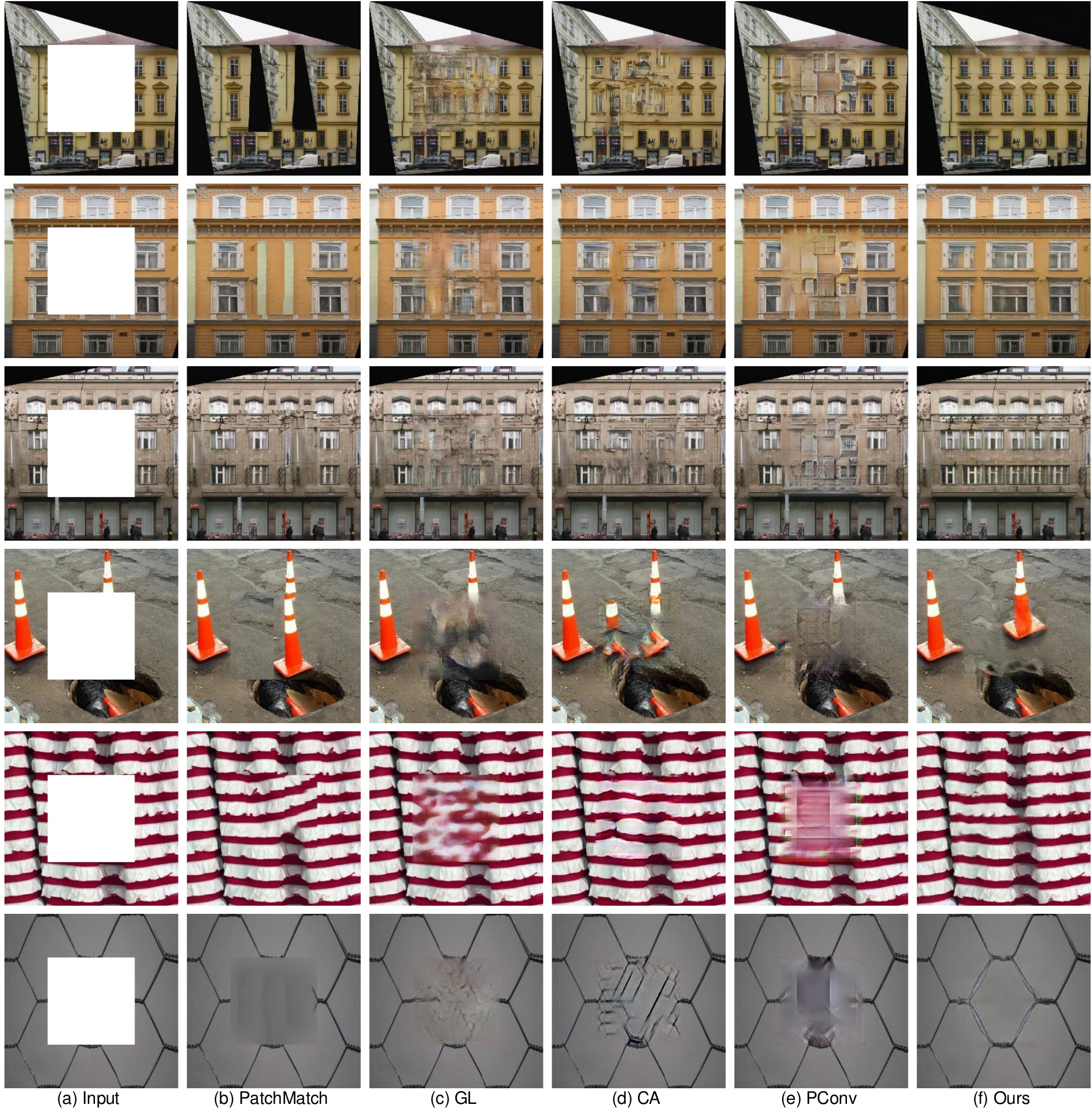}
	\end{center}
	\caption{Qualitative comparisons on the test images from Facade and DTD with square masks. In each row, we show from left to right the input, results from PatchMatch, GL, GntIpt, PConv and our model. Compared with baselines, our model is able to generate clear textures and structures that are consistent with context.~[Best viewed with zoom-in.]}
	\label{fig:facade_dtd_square}
\end{figure*}

\begin{figure*}
	\begin{center}
		\includegraphics[width=\linewidth]{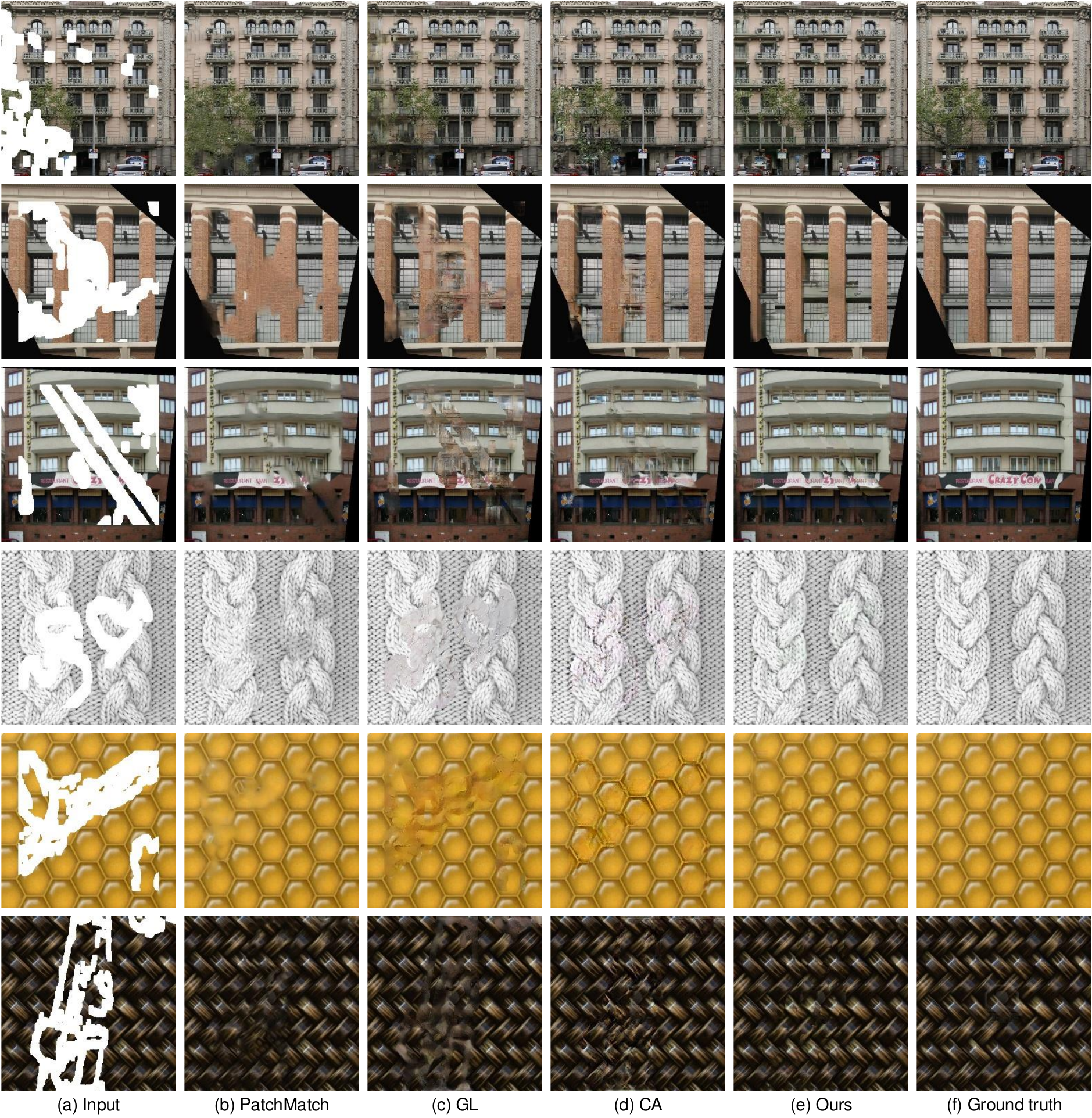}
	\end{center}
	\caption{Qualitative comparisons on the test images from Facade and DTD with irregular masks. In each row, we show from left to right the input, results from PatchMatch, GL, GntIpt, our model, and the ground truth. Compared with baselines, the semantic structures and texture patterns of the results are well-preserved by our model.~[Best viewed with zoom-in.]}
	\label{fig:facade_dtd_pconv}
\end{figure*}

\begin{figure*}
	\begin{center}
		\includegraphics[width=\linewidth]{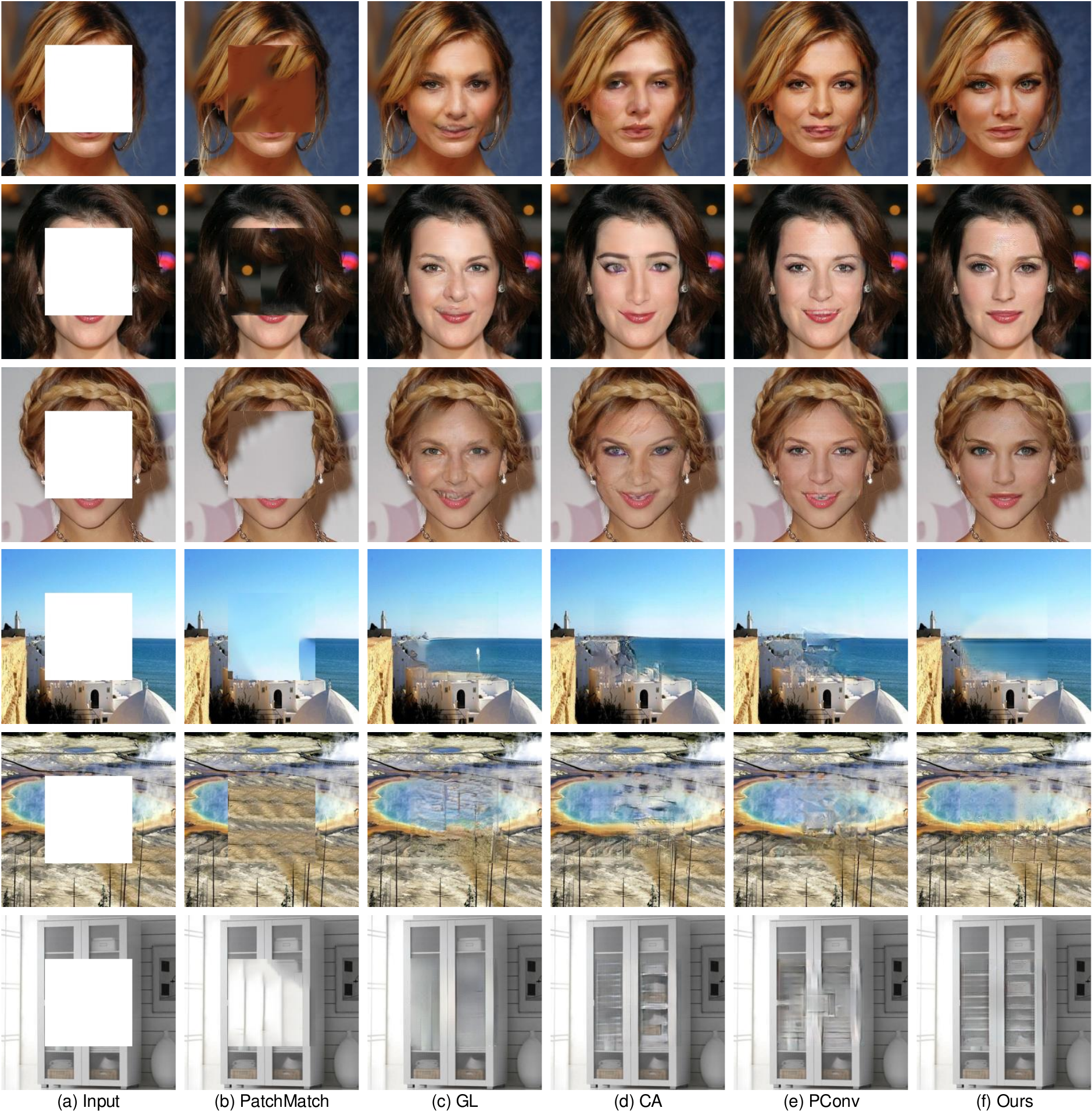}
	\end{center}
	\caption{Qualitative comparisons on test images from CELEBA-HQ and Places2 with square masks. In each row, we show from left to right the input, results from PatchMatch, GL, GntIpt, PConv and our model. Compared with baselines, our model is able to generate more natural results for images of human faces and natural scenes.~[Best viewed with zoom-in.]}
	\label{fig:celebahq_places2}
\end{figure*}

\begin{figure*}
	\begin{center}
		\includegraphics[width=\linewidth]{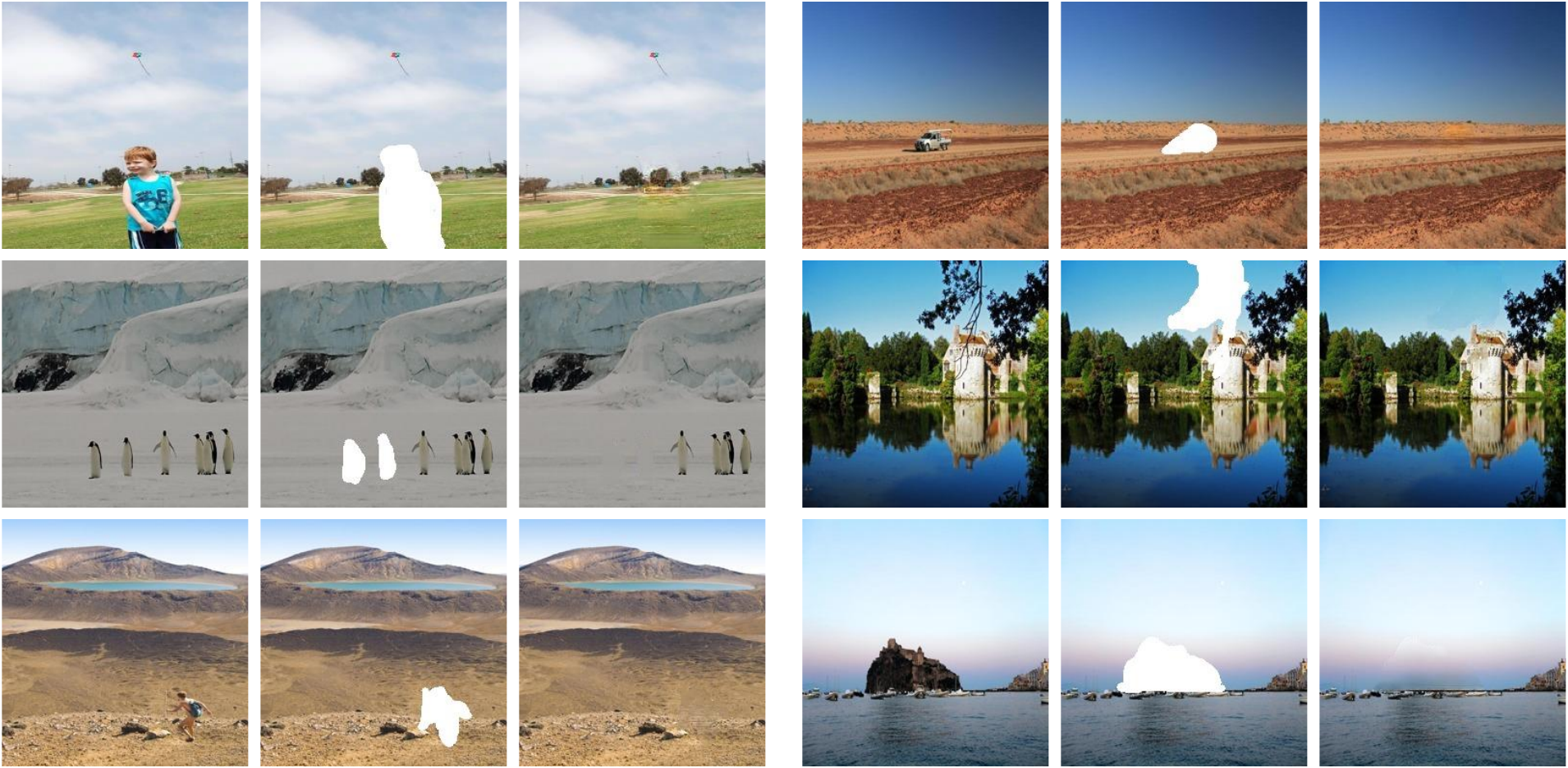}
	\end{center}
	\caption{Visual results for object removal on images of natural scenes. Our model is able to generate semantically-reasonable and visually-realistic results, which shows promising applications in user scenarios.~[Best viewed with zoom-in.]}
	\label{fig:removal}
\end{figure*}

In addition to Section 4, we show more qualitative comparisons on Facade~\cite{Tylecek13}, DTD~\cite{cimpoi14describing}, CELEBA-HQ~\cite{liu2015faceattributes} and Places2~\cite{zhou2018places} in Figure~\ref{fig:facade_dtd_square}~(SUPP.), Figure~\ref{fig:facade_dtd_pconv}~(SUPP.) and Figure~\ref{fig:celebahq_places2}~(SUPP.). More visual results for object removal on images of natural scenes are shown in Figure~\ref{fig:removal}~(SUPP.).

\begin{table}[!htb]
	\begin{center}
		\begin{tabular}{l}
			\hline
			\textbf{Input}: Image $\oplus$ Mask~($256\times256\times4$)\\ \hline
			$\phi^1$: Conv. (3, 3, 16), stride=1; LReLU;\\ \hline
			$\phi^2$: Conv. (3, 3, 32), stride=2; LReLU;\\ \hline
			$\phi^3$: Conv. (3, 3, 64), stride=2; LReLU;\\ \hline
			$\phi^4$: Conv. (3, 3, 128), stride=2; LReLU;\\ \hline
			$\phi^5$: Conv. (3, 3, 256), stride=2; LReLU;\\ \hline
			$\phi^6$: Conv. (3, 3, 256), stride=2; LReLU;\\ \hline
			$\phi^7$: Conv. (3, 3, 256), stride=2; ReLU;\\ \hline \hline
			$\psi^6$: ATN($\phi^6$, $\phi^7$);\\ \hline 
			$\psi^5$: ATN($\phi^5$, $\psi^6$);\\ \hline 
			$\psi^4$: ATN($\phi^4$, $\psi^5$);\\ \hline 
			$\psi^3$: ATN($\phi^3$, $\psi^4$);\\ \hline 
			$\psi^2$: ATN($\phi^2$, $\psi^3$);\\ \hline 
			$\psi^1$: ATN($\phi^1$, $\psi^2$);\\ \hline 
		\end{tabular}
	\end{center}
	\vspace{-5mm}
	\caption{The architecture of the pyramid-context encoder. $\phi^i$ denotes the feature map in the $i$-th layer defined in Section 3.1, $\psi^i$ denotes the reconstructed features by the ATN in the $i$-th layer defined in Section 3.1, and LReLU denotes leaky ReLU with the slope of 0.2.}
	\label{table:encoder}
\end{table}

\begin{table}[htbp]
	\vspace{-2.85cm}
	\begin{center}
		\begin{tabular}{l}
			\hline
			$\varphi^6$: DeConv. (3,3,256), stride=2; ReLU; $\oplus$ $\psi^6$ \\ \hline
			$\varphi^5$: DeConv. (3,3,256), stride=2; ReLU; $\oplus$ $\psi^5$ \\ \hline
			$\varphi^4$: DeConv. (3,3,128), stride=2; ReLU; $\oplus$ $\psi^4$ \\ \hline
			$\varphi^3$: DeConv. (3,3,64), stride=2;  ReLU; $\oplus$ $\psi^3$ \\ \hline
			$\varphi^2$: DeConv. (3,3,32), stride=2;  ReLU; $\oplus$ $\psi^2$ \\ \hline
			$\varphi^1$: DeConv. (3,3,16), stride=2;  ReLU; $\oplus$ $\psi^1$ \\ \hline \hline
			$output\_6$: Conv. (1,1,3), stride=1; clip; \\ \hline
			$output\_5$: Conv. (1,1,3), stride=1; clip; \\ \hline
			$output\_4$: Conv. (1,1,3), stride=1; clip; \\ \hline
			$output\_3$: Conv. (1,1,3), stride=1; clip; \\ \hline
			$output\_2$: Conv. (1,1,3), stride=1; clip; \\ \hline
			$output\_1$: Conv. (1,1,3), stride=1; clip; \\ \hline
		\end{tabular}
	\end{center}
	\vspace{-5mm}
	\caption{The architecture of the multi-scale decoder. $\oplus$ denotes feature concatenation, $\varphi^i$ denotes the feature maps in the decoder defined in Section 3.2, $output\_i$ denotes the predictions made by the decoder at each scale.}
	\label{table:decoder}
\end{table}

\begin{table}[!htbp]
	\vspace{-8cm}
	\begin{center}
		\begin{tabular}{l}
			\hline
			\textbf{Input}: Image~($256\times256\times3$)\\ \hline
			[layer 1]: SNConv. (5,5,64), stride=2; LReLU; \\ \hline
			[layer 2]: SNConv. (5,5,128), stride=2; LReLU; \\ \hline
			[layer 3]: SNConv. (5,5,256), stride=2; LReLU; \\ \hline
			[layer 4]: SNConv. (5,5,512), stride=2; LReLU; \\ \hline
			[layer 5]: SNConv. (5,5,1), stride=1; \\ \hline
		\end{tabular}
	\end{center}
	\vspace{-5mm}
	\caption{The architecture of the discriminator. \textit{SNConv.} denotes the convolutions with spectral normalization.}
	\label{table:D}
\end{table}

\end{document}